\documentclass[sigconf]{acmart}

\AtBeginDocument{%
  \providecommand\BibTeX{{%
    \normalfont B\kern-0.5em{\scshape i\kern-0.25em b}\kern-0.8em\TeX}}}

\copyrightyear{2021}
\acmYear{2021}
\setcopyright{acmcopyright}
\acmConference[KDD '21]{Proceedings of the 27th ACM SIGKDD Conference on Knowledge Discovery and Data Mining}{August 14--18, 2021}{Virtual Event, Singapore}
\acmBooktitle{Proceedings of the 27th ACM SIGKDD Conference on Knowledge Discovery and Data Mining (KDD '21), August 14--18, 2021, Virtual Event, Singapore}
\acmPrice{15.00}
\acmDOI{10.1145/3447548.3467160}
\acmISBN{978-1-4503-8332-5/21/08}

\acmSubmissionID{ads2768}

\usepackage{tabulary}
\usepackage{amsmath}
\usepackage{amsfonts}
\usepackage{algpseudocode}
\usepackage{algorithm}

\algnewcommand\algorithmicforeach{\textbf{for each}}
\algdef{S}[FOR]{ForEach}[1]{\algorithmicforeach\ #1\ \algorithmicdo}

\DeclareMathOperator*{\argmin}{arg\,min}

\begin{document}
\fancyhead{}

\title{SizeFlags: Reducing Size and Fit Related Returns in Fashion E-Commerce}



\author{Andrea Nestler}
\authornote{All authors contributed equally \\ $\ddag$~Corresponding Authors: {andrea.nestler@zalando.de}, {reza.shirvany@zalando.de} \\ $\dag$ Work done while at Zalando SE} \authornotemark[3]
\email{andrea.nestler@zalando.de}
\affiliation{
  \institution{Zalando SE}
  \city{Berlin}
  \country{Germany}
}

\author{Nour Karessli}
\authornotemark[1]
\email{nour.karessli@zalando.de}
\affiliation{
  \institution{Zalando SE}
  \city{Berlin}
  \country{Germany}
}

\author{Karl Hajjar}
\authornotemark[1] \authornotemark[2]
\email{karl.hajjar@polytechnique.edu}
\affiliation{
  \institution{Paris-Saclay University}
  \city{Paris}
  \country{France}
}

\author{Rodrigo Weffer}
\authornotemark[1]
\email{rodrigo.weffer@zalando.de}
\affiliation{
  \institution{Zalando SE}
  \city{Berlin}
  \country{Germany}
}

\author{Reza Shirvany}
\authornotemark[1] \authornotemark[3]
\email{reza.shirvany@zalando.de}
\affiliation{
  \institution{Zalando SE}
  \city{Berlin}
  \country{Germany}
}

\renewcommand{\shortauthors}{Nestler, et al.}


\begin{abstract}
\noindent E-commerce is growing at an unprecedented rate and the fashion industry has recently witnessed a noticeable shift in customers' order behaviour towards stronger online shopping. However, fashion articles ordered online do not always find their way to a customer's wardrobe. In fact, a large share of them end up being returned. Finding clothes that fit online is very challenging and accounts for one of the main drivers of increased return rates in fashion e-commerce. Size and fit related returns severely impact 1.~the customers experience and their dissatisfaction with online shopping, 2.~the environment through an increased carbon footprint, and 3.~the profitability of online fashion platforms. Due to poor fit, customers often end up returning articles that they like but do not fit them, which they have to re-order in a different size. To tackle this issue we introduce {SizeFlags}, a probabilistic Bayesian model based on weakly annotated large-scale data from customers. Leveraging the advantages of the Bayesian framework, we extend our model to successfully integrate rich priors from human experts feedback and computer vision intelligence. Through extensive experimentation, large-scale A/B testing and continuous evaluation of the model in production, we demonstrate the strong impact of the proposed approach in robustly reducing size-related returns in online fashion over~14~countries.
\end{abstract}





\maketitle

\section{Introduction}
\label{sec:introduction}
\noindent Article returns are critical to any retail industry where customers return those articles which they find unsatisfactory for various reasons. Customers in turn receive a full or a partial refund for their returns according to the retailer's specific return policy. Over the past years, customer returns have been growing at a significant rate reaching up to $50\%$ increase year over year for certain categories~\cite{Stalk2006,Choi2016}. Article return rates vary greatly among retail industries, categories, brands, and distribution channels~\cite{Langley2008, Ofek2011, Barry2000}. However, the highest return rates are in online fashion apparel with $25-40\%$ overall return rates that can reach up to $75\%$ for specific categories and brands~\cite{Barry2000, Mostard2006}. The underlying reason is that fashion apparel involves complex factors such as size, fit, color, style, taste and unquantifiable factors such as ``it's not me''~\cite{Barry2000}. Among these reasons, poor size and fit is cited as the number one factor in online fashion returns~\cite{Ratcliff2014}.  
On one hand, physical examination of fashion articles to assess their size, fit, fabric, design and pairing with other fashion articles is crucial for customers in their order decisions~\cite{Ha2004}. On the other hand, in online fashion customers order garments and shoes without the possibility of trying them on and this crucial sensory and visual feedback is delayed to the unboxing experience. The absence of ``feel and touch'' experience therefore leads to major uncertainties in the buying process and to the hurdle of returning articles. As such, many customers either hesitate to place an order, or opt for various strategies for reducing the uncertainty, such as ordering multiple sizes or colors of the same article and then returning those that did not match their criteria - even more so for fashion categories and brands they are less familiar with. 

Escalating the problem for size and fit, fashion articles suffer from significant sizing variations~\cite{Ashdown2007} due to: different sizing systems (Alpha, Numeric, Confection); coarse definition of size systems (S, M, L for garments); country conventions (EU, FR, IT, UK); different specifications for the same size according to the brand; \textit{vanity sizing}~\cite{Weidner2010} where brands deliberately adapt their nominal sizes to target segment of customers based on age, sportiness, etc.; and different ways of converting a local size system to another. Moreover, customers' perception of size and fit for their body remains highly personal and subjective which influences what the right size is for each customer. The combination of the aforementioned factors leaves the customers alone to face a highly challenging problem of determining the right size and fit during their order journey. The problem of size and fit has a major impact on the environment, and the profitability of online fashion retailers. A major issue in sustainability is the substantial carbon footprint incurred in the logistic process of returning articles~\cite{Velazquez}. Online fashion retailers are strongly inclined to offer lenient return policies to lower customer perceptions of uncertainty~\cite{Yu19} which inherently leads to an increase in returns. As customer expectations around fit and sustainability evolve, online fashion retailers have to quickly adapt and find innovative techniques to conclusively reduce the high size and fit related return rates. In recent years, an emerging body of multi-disciplinary research has been proposed with the aim of enabling customers to find something that fits them from the first time as discussed in the related work section~\cite{walsh14, Zhang,Diggins2016ARC,Cullinane, guigoures18,saboor19,julia20,Lefakis2020,Hajjar2020, embed-for-reco-asos, abdulla2017,sembium2017,Sembium2018,Du2019,Vecchi2015LookingFT,Sajan19,sizenet,hsiao20,han2018viton,Guan12drape:dressing,Dong_2019_ICCV}. 

In this work we introduce a powerful approach, called SizeFlags, to provide customer agnostic size advice with strong impact on decreasing size and fit related returns in fashion e-commerce. This approach naturally benefits from the advantages of Bayesian methods - modeling uncertainty and the use of priors. Considering the lack of size and fit expert labeled data, we rely on large-scale weakly annotated data from the returns process. More specifically, first we leverage the crowd’s subjective and noisy return reason feedback
(which is highly influenced by individuals perception of size and fit) as an input signal to determine the fit of an article. We thus construct a binomial model based on the return behaviour of articles and then extend it to a fully Bayesian setting which integrates rich priors from both human expert feedback and computer vision intelligence. We present multiple versions of our model including the baseline algorithm launched for textile and shoe categories in 2017, and the fully Bayesian SizeFlags algorithm launched in early 2020, currently in use for millions of articles ordered over 14 countries.\\ \textbf{The contributions of this work are:} (1) We introduce the Baye\-sian SizeFlags framework to determine the fit behaviour of articles based on subjective noisy and anonymized return data from customers,
(2) We integrate, for the first time to our best knowledge, human fashion expert feedback as well as computer vision based cues as rich priors for tackling the size advice problem and demonstrate these priors strongly contribute in reducing size and fit related returns - by addressing the challenging cold-start problem for the thousands of new articles appearing on shopping platforms every day and (3) we demonstrate with extensive experimental results obtained through A/B testing and continuous in-production evaluation, how each part of our model contributes to reducing size and fit related returns.

We note that, evaluating the impact of size and fit recommendations remains highly challenging in the literature due to multiple underlying factors specific to this problem space; on one hand the ``true'' size of a customer is often unknown, remains subjective for each customer, and can vary greatly by external factors including life changing events impacting a customer's physical body, and/or mindset around what fits best. On the other hand, looking at the problem from the article side, the right size for a customer is not a unique quantity and varies greatly both within and across hundreds of brands, in different sizing systems, in different countries, and for different fashion categories.  Therefore, within this work along side introducing our novel size advice approach, we also make a substantial attempt (a first to the best of our knowledge) to establish the state-of-the-art baseline, and rigorously assess the impact of size recommendations with respect to \textit{reducing size-related returns} in online fashion.



\vspace*{-0.4\baselineskip}

\section{Related Work}
\label{sec:related-work}
\noindent The customers' struggle of finding the right size while shopping online is a well known challenge in the fashion industry~\cite{Ratcliff2014,Ashdown2007,Ha2004,Weidner2010,walsh14, Zhang,Diggins2016ARC,Cullinane, guigoures18,saboor19,julia20,Lefakis2020,Hajjar2020, embed-for-reco-asos, abdulla2017,sembium2017,Sembium2018,Du2019,Vecchi2015LookingFT,Sajan19,sizenet,hsiao20,han2018viton,Guan12drape:dressing,Dong_2019_ICCV}. ~\cite{Diggins2016ARC, Cullinane} address issues related to customer returns in online fashion retailing and discuss their implications while~\cite{walsh14} studies customer-based preventive article return management. \cite{Zhang} suggests a crowdsourcing approach where customers get suggestions on garments based on matching the existing articles in their wardrobes to others in the community.  With the aim of unifying the different sizes across size systems and brands,~\cite{Du2019} propose a method to automate size normalization to a common latent space through the use of articles ordered data, generating mapping of any article size to a common space in which sizes can be better compared. More recently, there has been emerging research addressing the problem of personalized size recommendation for online fashion retailers \cite{guigoures18,saboor19,julia20,Lefakis2020,Hajjar2020, embed-for-reco-asos, abdulla2017,sembium2017,Sembium2018}. Given the order history of a customer (or personal customer data such as age, weight, height, etc.), these methods predict for that customer which size of an article would fit best. Such size recommendation systems personalized to the customer have proved high value in supporting customer decision and enhancing their experience on the platform with respect to size and fit with strong impact on the customer conversion. However,~\cite{Vecchi2015LookingFT} shows that most customers who receive a correct size recommendation would not buy the size recommended due to their individual fit preferences. Additionally, those methods have to explicitly deal with the potentiality of having multiple customers, and thus multiple size and fit preferences behind a single account so to present the right recommendation to its target customer behind the account.  However, it should be stated here that to the best of our knowledge, the positive impact of such systems on reducing size-related returns has not yet been demonstrated. We further develop on this matter in~\autoref{sec:experiments} by implementing and A/B testing recent well-known personalized recommendations described in~\cite{guigoures18} and assess their potential impact on size-related returns.

From a radically different angle, other methods~\cite{han2018viton,Guan12drape:dressing,Dong_2019_ICCV} try to estimate a customer's body shape using computer vision (2D or 3D modelling). Although such approaches can efficiently create realistic looking avatars, the claimed virtual fitting experience cannot accurately portray the actual garment fit on the individual complex body. What is more, such virtual try-on methods require personal data from customers such as images in tight clothing, age, gender, weight, height, etc. From a different angle, ~\cite{hsiao20} suggests learning body-aware embeddings to recommend clothing which complements a specific body type using mined attributes with visual features for clothing, and estimated body shapes. However the customer body shape estimation suffers from the same personal data requirement as the works mentioned above. Relaxing the constraint on having personal customer data,~\cite{sizenet} proposes to focus on articles only and suggests using article images to predict how likely a given article is to have a general size issue.

Looking into reducing returns, \cite{Sajan19} proposes a method to predict the probability that a customer will return a specific article before the order is placed. Although, the deep neural network architecture introduced is able to capture some latent size and fit information about a customer, the used article features are not size and fit specific (even though the authors acknowledge that a major part of the returns are due to poor size or fit). When an order is found likely to be returned, the suggested approach tries to limit that order with candidate mechanisms including placing more burden on the customer by prohibiting a return, introducing an additional fee, or stopping the order altogether. 

In this paper, we aim to both support customers in their size and fit decisions—without imposing any burden of personal data or limitations on their article orders—and reduce size and fit related returns. To that end, we present a Bayesian model which uses size-related return data from customers to learn which articles would fit normally and which would exhibit a size issue. Unlike previous work~\cite{guigoures18,saboor19,julia20,Lefakis2020,Hajjar2020, embed-for-reco-asos, abdulla2017,sembium2017,Sembium2018}, our approach extensively leverages the size-related return rates of articles to greatly focus on modeling articles sizing behaviour. Although the (weakly annotated and subjective) return data from customers is leveraged in the model, the latter is agnostic to the specific customer who places the order and thus by design does not provide a customer with a personalized size recommendation; instead it informs them about article specific sizing characteristics. Thanks to this strategy we create an algorithmically driven customer experience that presents a very low cognitive load by purposefully focusing our approach towards the articles themselves and the underlying manufacturing and brand characteristics, in contrast to personalized size recommendations~\cite{guigoures18,saboor19,julia20,Lefakis2020,Hajjar2020, embed-for-reco-asos, abdulla2017,sembium2017,Sembium2018} which are emotionally engaging for the customers by trying to convey to them ``what their size is''—strongly increasing the chances of a correct recommendation being dismissed altogether~\cite{Vecchi2015LookingFT}—our approach rather focuses on the articles themselves and supports customers to make an informed decision on which size to order given the inferred size and fit characteristics of a given article. This in turn, gives customers the flexibility to adapt their choice based on their fit preference instead of having to cope with a rigid recommendation engine telling them what their size is. Therefore, in contrast to~\cite{guigoures18, saboor19, julia20} which evaluate the performance of their methods solely based on the accuracy of the personalized size recommendations (i.e. whether the system is good at inferring a customers' size), in this work we go a huge step forward and evaluate the impact of our system on the end-to-end size and fit journey; beyond onsite and unboxing stages and at the size-related return level. Detecting size issues on an individual article level also alleviates other challenges like having multiple users behind one account and having different size systems which~\cite{sembium2017, Sembium2018, guigoures18, embed-for-reco-asos, abdulla2017,Sajan19} have to explicitly deal with.

\vspace*{-0.4\baselineskip}

\section{Approach}
\label{sec:approach} 


\begin{figure*}
    \includegraphics[width=.69\textwidth]{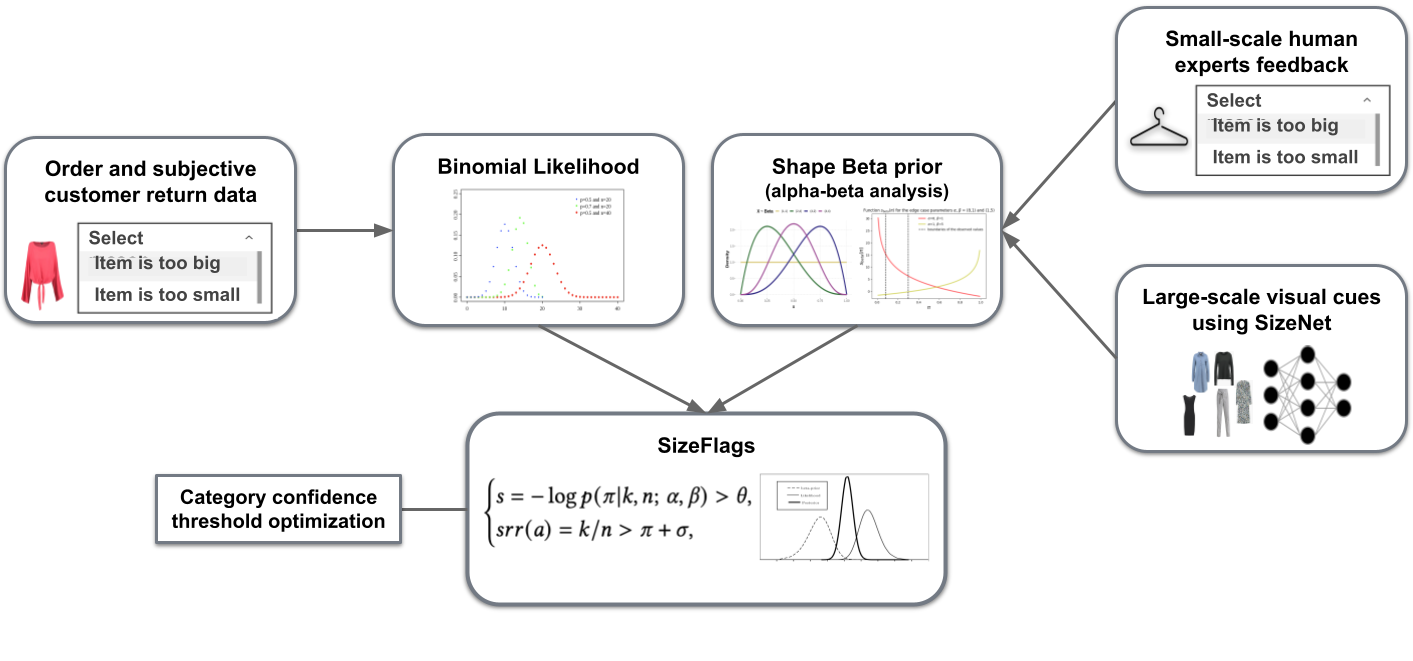}    
    \vspace{-0.8cm}
    \caption{High level overview of the SizeFlags approach.}
    \Description{High level overview of the SizeFlags approach, a probabilistic Bayesian model based on weakly annotated large-scale data from customers and integrates rich priors from human experts feedback and computer vision intelligence.}
\vspace*{-0.9\baselineskip}
    \label{fig:architectural_diagram}
\end{figure*}

\noindent The SizeFlags model we present is fully Bayesian and aims at modeling the behaviour of the size-related returns of an article using some prior information about the article as well as orders and returns data.~\autoref{fig:architectural_diagram} shows the high level overview of the approach. We use a Binomial distribution for the likelihood and a Beta distribution for the prior. As we will discuss below, prior information can come from various sources of different complexity to help shape the prior Beta distribution. Among the more sophisticated ways, we will discuss obtaining prior information about the fit of an article before observing any return using advanced computer vision deep learning models, such as~\cite{sizenet}, to process the images of the article. 

Let us first simplify the complex problem of determining the size and fit of fashion articles into a $3$-class classification problem \{no size issue, too small, too big\} as in~\cite{sembium2017,Sembium2018}, where we have summarized the possible size issues of articles into two categories: `too small' (indicating the article runs smaller than usual e.g. too short, shoulder too tight, etc) and `too big' (indicating the article runs bigger than usual e.g. too loose, sleeves too long, etc). To determine whether an article has a size issue or not we use a Bayesian model based on a binomial likelihood.  In practice, we use two Bayesian models to predict separately if an article is `too big' or not, and if it's `too small' or not.
In this context, an article is classified as `no size issue' if it is tagged by neither of the two Bayesian models. Although the events `too big' and `too small' are not independent for a given article, we choose to model them in this way for simplicity and leave the extension to a unified multinomial model for future work. To make things easier for the reader, in all that follows we will present a single model to tag whether an article has a size issue or not, although in practice we use two separate models to tag `too big' versus `not too big' and `too small' versus `not too small'.

In~\autoref{subsec:binomial} we introduce the binomial likelihood model and explain in~\autoref{sec:flag-conditions} how this model can be used to predict size issues (raise a size issue flag). In~\autoref{sec:limitations} we discuss the cold-start problem related to having new articles on the platform, and describe in~\autoref{sec:full-model} how the incorporation of priors, making the model fully Bayesian, can help tackle this problem. 

%

\subsection{The Binomial Likelihood}
\label{subsec:binomial}
We start by formalizing the problem, given an article $a \in \mathcal{C}$ that belongs to a fashion category $\mathcal{C} \subset \mathcal{C}_{all}$ (\textit{e.g.} jeans, shirts, shoes, etc.). We define its observed size-related return rate $srr(a)$ as the ratio of the article returns for reasons `too big' (or similarly `too small') $k$ out of the number of the total article orders $n$. We denote the true size-related return rate of an article as $srr_{true}(a)$. This value is unknown, but we know that the higher the number of article orders $n$, the more confident we are that $srr(a)$ is close to $srr_{true}(a)$:   $srr_{true}(a)=\lim_{n\rightarrow\infty} srr(a)$. Similarly to~\cite{sizenet}, we consider two main elements:

\noindent\textbf{(1) Sales period.} Depending on the period during which an article is sold, customers' return behaviour and article's ratio of size issues in a category might vary. 

\noindent\textbf{(2) Fashion category.} Each fashion category $\mathcal{C}$ poses different size and fit challenges. Therefore, we compute the mean 
\vspace*{-0.9\baselineskip}

$$\pi= \text{mean}(\{srr(a)\}_{a\in \mathcal{C}})$$ 
\vspace*{-0.9\baselineskip}

and the standard deviation 
\vspace*{-0.9\baselineskip}

$$\sigma= \text{std}(\{srr(a)\}_{a\in \mathcal{C}})$$ 
\vspace*{-0.9\baselineskip}

of size-related return rate of all articles in the category.

The article $a$ could be considered to have a size issue if $srr_{true}(a)$ $> \pi+\sigma$
when the return rates are calculated over the same articles ordered period of time. Thus, as in~\cite{sizenet}, we use a binomial law to assess the confidence in the class prediction. The probability of observing $k$ size-related returns over a total of $n$ orders for that article $a$ is modelled by the binomial likelihood
\begin{equation}\label{eq:binomial}
    p(k\vert n , \pi) = \binom{n}{k} \pi^k (1-\pi)^{n-k}.
\end{equation}
This likelihood is maximized when the ratio of $k$ over $n$ is equal to $\pi$. In other words, when $k$ is the expected number of size-related returns when drawing $n$ samples from a Bernoulli distribution of parameter $\pi$. The estimator becomes more confident whether the $k$ returns were actually drawn with probability $\pi$ as more article orders are observed. As $n$ grows larger, the likelihood will be close to $1$ if the ratio $k/n$ is close to $\pi$, and close to $0$ if it diverges from it. On the other hand, for low values of $n$, the estimator is more uncertain and tends to weigh uniformly all possible values of $k$. We use the score $s$ defined in~\cite{sizenet}, associated to the observation of $k$ returns out of $n$ article orders given $\pi$ based on the negative log-likelihood:
\begin{equation}
    s = -\log\,{p(k\vert n , \pi)}.
    \label{eq:est_score}
\end{equation}

This score $s$ is a positive number and for large values of $n$, the score~\eqref{eq:est_score} scales linearly with respect to $n$ with the divergence rate of the article size-related return rate probability distribution in comparison to its category. For a detailed analysis of the score behavior, we refer the reader to~\cite{sizenet}.

\subsection{Raising a size issue flag}\label{sec:flag-conditions}
%
%
For a given article $a$ for which we have observed $k$ returns out of $n$ article orders, and given $\pi$, we can query the binomial likelihood $p(k,n|\pi)$ defined in~\eqref{eq:binomial}, or alternatively its negative log defined in~\eqref{eq:est_score}, to know if this article has a size issue or not. 
A high likelihood means that the observed $srr(a)=k/n\approx\pi$ likely has a ``normal'' sizing behaviour. 
On the other hand, a low likelihood could mean one of two things: either $srr(a)$ is too high compared to $\pi$ ($srr(a)\gg \pi$) and has therefore a size issue, or it is too low ($srr(a)\ll \pi$) then the article is behaving better than average in its category. Ideally, if we had an infinite number (or more realistically a very large number) of orders $n$ for a given article $a$, then the observed $srr(a)$ would tend to the ``true'' return rate $srr_{true}(a)$ of the article $a$, and we would need only compare this value to 
$\pi$ to decide if the article indeed has a size issue or not. 
Then we could raise a ``size issue'' flag if%
\begin{equation}\label{eq:flag-condition}
    srr_{true}(a) \geq \pi + \sigma,
\end{equation}
\noindent where $\sigma$ is the standard deviation of $srr(a)$ over the category $\mathcal{C}$. However, in a realistic setting, using the condition of~\eqref{eq:flag-condition} is not enough to decide whether an article has a size issue or not. Indeed, for an article with a low number of orders (say $n=10$), the condition could be verified but we do not have enough article orders to make sure that $srr(a)$ is actually close to $srr_{true}(a)$. This is where the binomial likelihood comes into play, as it is able to account for that uncertainty. As mentioned in ~\autoref{subsec:binomial}, this likelihood tends to be relatively flat for low values of $n$, resulting in probabilities which are not too low for any value of $k$. We set a very low error bound $\varepsilon >0$ as an upper bound on $p(k,n|\pi) < \varepsilon$, and use this in combination with the condition of~\eqref{eq:flag-condition} to determine whether an article has a size issue or not. The condition on the likelihood will ensure that the binomial model is sure enough that $srr(a)$ is ``abnormal'' compared to what is observed over the whole category $\mathcal{C}$, and that is not just due to lack of article orders. The condition of~\eqref{eq:flag-condition} will tell us in turn that the return rate $srr(a)$ is abnormally high, resulting in a size issue flag, and not abnormally low compared to the rest of the category (which would mean a very well fitting article). 

Using $\varepsilon$ as an upper bound on the likelihood translates into a lower bound threshold $\theta = - \log(\varepsilon) > 0$ on the score $s$ from~\eqref{eq:est_score}. Combining the two conditions described above into a single statement leads to the following joint condition for raising a size issue flag: 
\begin{align}\label{eq:combined-conditions-flag}
    \begin{cases}
      s = - \log p(k\vert n, \pi) \geq \theta,\\
      srr(a)=k/n \geq \pi + \sigma.
    \end{cases}
\end{align}

\subsection{Parameter setting and cold-start problem}\label{sec:limitations}


With the two joint conditions in~\eqref{eq:combined-conditions-flag}, the question is now how to set the threshold $\theta=-\log(\varepsilon)$ that ensures high confidence. 
Here are 2 ways to set an appropriate parameter $\theta$ for the algorithm:
\begin{enumerate}
    \item If the goal is to identify the $x$ most problematic articles (e.g. $5 \%$ of a category $\mathcal{C}$), then we can set the steering parameter $\theta$ by calibrating this threshold on past data.
    \item If we want to raise a sizing flag with high certainty, then $\varepsilon>0$ must be chosen very small. In this case, $\varepsilon$ can be set for example equal to machine epsilon $\varepsilon_{mach}$ (e.g. in single precision $\varepsilon_{mach} = 2^{-23} \approx 1.19 e^{-07}$).
\end{enumerate}

In calibrating this threshold, we run through the risk of taking too long until the estimator achieve the required confidence level. This risk is particularly relevant in fashion e-commerce, where articles generally have a short lifetime on the platform and return data is delayed due to logistic constraints. Therefore, in the discussed approach, the article might run out of stock before a sufficient amount of orders and returns is observed. This directly leads to customers dissatisfaction and increased environmental and financial costs as more returns directly translates to a higher carbon footprint and a higher logistic costs. Moreover, with this high confidence bar, problematic articles with low number of orders might never get flagged during their whole lifetime on the platform. This means that customers are not informed about many articles that have size issues. 



\subsection{Threshold optimization}\label{sec:threshold_optimization}

Setting a conservative threshold $\theta$ makes the algorithm more robust to noise and increases the confidence. This hinders the algorithm's performance as too high numbers of article orders and returns ($n$ and $k$) are necessary to raise a size issue flag, interfering with the main goal of the algorithm.
To tackle this cold-start problem, we propose to choose a more conservative $\varepsilon_{\min}$ in the beginning to get stable and accurate sizing flags first. Based on this $\varepsilon_{\min}$ and $\theta_{\max}=-\log(\varepsilon_{\min})$ selected as a starting point, a better threshold $\theta^{\star}$ for ~\eqref{eq:combined-conditions-flag} can be determined by solving an optimization problem. Taking into consideration that different fashion categories exhibit different sizing challenges and return behaviour, consequently we search for optimal threshold per category.

We show now how to get an optimized threshold $\theta^\star$ $=\theta^\star(\mathcal{C})$ using historical data in order to obtain almost the same size issue flags as with $\theta_{\max}$, only much faster. We select $\theta^\star>0$ according to the following problem:
\begin{align}\label{threshold_opti}
    &\theta^\star = \argmin_{\theta \in (0, \, \theta_{\text{max}}]} \theta \ \ \text{ subject to } \ \ \frac{N(\mathcal{C},\theta)-S( \mathcal{C}, \theta)}{N( \mathcal{C}, \theta)} \leq \varepsilon_1, \\ \nonumber
    &\frac{N(\mathcal{C}, \theta)-N(\mathcal{C}, \theta_{\max})}{N(\mathcal{C}, \theta_{\max})} \leq \varepsilon_2, \ \ \ \frac{N(\mathcal{C}, \theta)-S(\mathcal{C}, \theta)}{N(\mathcal{C}, \theta_{\max})-S(\mathcal{C}, \theta_{\max})} \leq \varepsilon_3.
\end{align}
\\
We denote by $N(\mathcal{C}, \theta)$ the number of articles $a\in \mathcal{C}$ flagged by the algorithm with parameter $\theta$ and by $S(\mathcal{C}, \theta)$ (the stability) the number of flags raised whose value do not change across the time period considered. 
Thus, the difference $N(\mathcal{C}, \theta)-S(\mathcal{C}, \theta)$ can be seen as the number of unstable flags. The second constraint restricts the number of additional flags compared to the baseline.
The third constraint translates that the ratio of unstable flags with parameter $\theta$ compared to the same ratio with the baseline $\theta=\theta_{\text{max}}$ must be restricted by~$\varepsilon_3$. Parameters~$\varepsilon_i$ are thus values set to ensure the flags raised using a threshold $\theta$ are similar in number and quality to those raised by baseline. Due to the monotony behavior of the constraints, $\varepsilon_1, \varepsilon_2\in(0,1)$ and $\varepsilon_3\geq 0$. 

In order to be able to reproduce our results, we would like to point out that we have solved the optimization problem (\ref{threshold_opti}) by first discretizing equidistantly in $\theta\in(0,\theta_{\max})$ and then choosing $\theta^{\star}$ with the smallest target-functional. We have set $(\varepsilon_1, \varepsilon_2, \varepsilon_3) = (0.2, 0.05, 1.5)$ in order to generate sufficiently stable results.

\subsection{Full Bayesian Model with priors}\label{sec:full-model}

To overcome the limitation (cold-start problem) exposed in ~\autoref{sec:limitations}, we propose the addition of a prior to our binomial likelihood model in order to introduce ``expert information'' to shape the binomial distribution even before any order is observed for an article. The expert information can come from different sources; we focus on Human expert feedback from fashion models and Sizing information extracted from visual cues of articles (SizeNet).
%
We use the Beta law as a prior distribution on $srr(a)$ as it is the conjugate distribution to the Binomial law. In this full Bayesian setting, we model each individual article as having a true, but unknown size-related return rate. We choose to model this return rate as a random variable $R$ following a Beta distribution $R \sim Beta(\alpha, \beta)$. Given a value of $R=r$ for an article and $n$ orders , the likelihood of observing $k$ returns for this article follows a Binomial distribution of parameter $r , \ $ $p_{binom}(k | n, r) = \binom{n}{k} r^k (1-r)^{n-k}$. The density function of the Beta distribution with parameters $\alpha, \beta > 0$ is given by $p_{\text{beta}}(r | \alpha, \beta) = B(\alpha, \beta)^{-1} r^{\alpha-1} \left (1-r \right )^{\beta -1}$ where $B(\alpha, \beta)$ is the value of the Beta function at $(\alpha, \beta)$. In this Bayesian setting, the full joint likelihood over variables $k, r$ is given by 
\begin{eqnarray}\nonumber
p(k, r | n, \alpha, \beta) &=& p_{\text{binom}}(k| n, r) p_{\text{beta}}(r | \alpha, \beta).
\end{eqnarray}

The Binomial law and the Beta law being conjugates, the posterior distribution $$p(r |k,n;\, \alpha, \beta) = B(k+\alpha, n-k+\beta)^{-1}r^{k+\alpha-1} \left (1-r \right )^{n-k+\beta -1}$$ over the true return rate $r$ of an article is also a Beta distribution of parameters $(k+\alpha-1,\, n-k+\beta-1)$.

\noindent\textbf{Raising a size issue flag:} In this new Bayesian setting, we replace the condition on the Binomial likelihood being small enough by a condition on a point estimate of the posterior distribution. The new joint conditions for raising a size issue flag now become:
\begin{align}\label{eq:combined-conditions-flag-posterior}
    \begin{cases}
      s_{\text{posterior}} = - \log p(\pi| k,n; \, \alpha, \beta) \geq \theta ,\\
      srr(a) = k/n  \geq \pi + \sigma,
    \end{cases}
\end{align}
\noindent where $\theta=\theta^\star$ can be obtained using the threshold optimization in ~\autoref{sec:threshold_optimization} and $\alpha$ and $\beta$ are article-specific parameters which are set using prior information about articles. Using this posterior distribution which incorporates expert knowledge through the prior distribution helps raise flags faster as discussed below.


\noindent\textbf{Prior distribution analysis:} In our modeling, the values of $\alpha$ and $\beta$ are defined as $\geq 1$ so that the prior distribution is not skewed around the extreme values $0$ and $1$. Since for $\alpha=\beta=1$ the prior represents a uniform distribution, we choose $\alpha,\beta \geq 1$ with at least one of them $>1$. $p_{\text{beta}}(r | \alpha, \beta)$ therefore becomes a unimodal distribution.


For the first condition of~\eqref{eq:combined-conditions-flag-posterior} it is essential that the parameters $\alpha$ and $\beta$ are reasonably restricted by $\alpha\in[1,\alpha_{\max}]$ and  $\beta\in[1,\beta_{\max}]$ where the upper boundaries strongly depend on the threshold $\theta$. Here $\alpha_{\max}$ and $\beta_{\max}$ may not be chosen too large to ensure a natural balance between $p_{\text{binom}}$ and $p_{\text{beta}}$ in condition~\eqref{eq:combined-conditions-flag-posterior}: raising a flag should not rely too strongly on the prior information only through $p_{\text{beta}}$. The prior plays a big role especially at the beginning when we do not have any sales and return data, i.e. n = 0 and k = 0. Therefore the optimization of $\alpha_{\max}$ and $\beta_{\max}$ is modeled for this edge case. With $p_0(\pi,\alpha,\beta) = -\log p(\pi| 0,0; \, \alpha, \beta)$ we impose the following conditions to get $\alpha_{\max}$ and $\beta_{\max}$:
\setlength{\arraycolsep}{0.3em}
\begin{eqnarray}\label{eq:alpha_beta_max_condition}
\begin{cases}
\begin{array}{ccl}
\alpha_{\max} &=& \argmin_{\alpha>1}\left|\max_{\pi\in\Pi} \big\{p_0(\pi,\alpha,1)\big\} - \theta\,\right|,\\[0.6em] 
\beta_{\max} &=& \argmin_{\beta>1}\left|\min_{\pi\in\Pi} \big\{p_0(\pi,1,\beta)\big\} + 1\,\right|.\\
\end{array}
\end{cases}
\end{eqnarray}
\\
\vspace*{-1.3\baselineskip}

\noindent The inner optimization problems $\pi_{\alpha}^\star = \arg\max_{\Pi}\{p_0(\pi,\alpha,1)\}$ and $\pi_{\beta}^\star = \arg\min_{\Pi}\{p_0(\pi,1,\beta)\}$ ensures that $p_0(\pi_{\alpha}^\star,\alpha_{\max},1) = \delta_{\alpha}\geq p_0(\pi,\alpha_{\max},1)$ and $p_0(\pi_{\beta}^\star,1,\beta_{\max}) = \delta_{\beta}\leq p_0(\pi,1,\beta_{\max})\,\,\forall\pi\in\Pi.$
By optimizing $\alpha$ to achieve $\theta$ at $\pi_{\alpha}^\star$, we ensure that $p_0$ for all other $\pi\in\Pi$ will be lower than $\delta_{\alpha}\approx\theta$. For $\beta$ it's the opposite: By optimizing $\beta$ to achieve $-1$ at $\pi_{\beta}^\star$, we ensure that $p_0$ for all other $\pi\in\Pi$ will be greater than $\delta_{\beta}\approx -1$.


One challenge in solving the optimization problems (\ref{eq:alpha_beta_max_condition}) is to find
a suitable choice of the fixed parameters $\theta > 0$ and $\Pi\subset[0,1]$:
In~\autoref{sec:limitations} we described how a suitable $\theta$ can be chosen.

The interval $\Pi=\Pi(\mathcal{C})$ depends on $\mathcal{C}$ and specifies the area in which the actual parameter $\pi$ is very likely to be located. $\pi$ can change over time because the category can evolve, so an interval is required here, in which $\pi$ usually is. Let's define fixed initial parameters $\mathcal{C}^{init}$ as the initial state of the category, $\pi^{init}=srr(\mathcal{C}^{init})$ and $\sigma^{init} =  \text{std}(\{srr(a)\}_{a\in \mathcal{C}^{init}})$ then we define
\vspace*{-0.7\baselineskip}

$$\Pi = [\pi^{init}-\sigma^{init}, \pi^{init}+\sigma^{init}].$$
\vspace*{-0.9\baselineskip}

\begin{figure}[ht]
    \centering
    \includegraphics[width=0.7\linewidth]{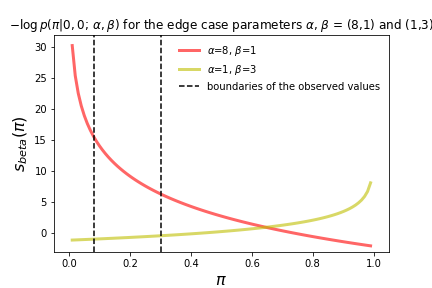}
    \vspace{-0.55cm}
    \caption{Prior information for $\alpha_{\max} = 8$ and $\beta_{\max} = 3$.}
    \Description{Prior distribution analysis: prior score for $\alpha_{\max} = 8$ and $\beta_{\max} = 3$.}
    \vspace*{-0.9\baselineskip}
    \label{fig:alpha_beta_edge_cases} 
\end{figure}

Exemplification: Assuming that $\Pi=[0.08,0.3]$ and $\theta=15$. Under the condition that both $\alpha\geq 1$ and $\beta\geq 1$ must be integers, we get $\alpha_{\max} = 8$ and $\beta_{\max} = 3$ as solution of the optimization problem (\ref{eq:alpha_beta_max_condition}). As you can see in~\autoref{fig:alpha_beta_edge_cases}, the functions $s_{\text{beta}}(\pi, 8,1)$ and $s_{\text{beta}}(\pi, 1,3)$ form an upper and lower boundary for $s_{\text{beta}}(\pi, \alpha,\beta)$ $\forall~ \alpha\in[1,8]$ and $\beta\in[1,3]$. All values $\alpha\in[1,8]$ and $\beta\in[1,3]$ can now be used for the algorithm, depending on the prior information of an article $a\in \mathcal{C}$.\vspace{-0.5cm}

\subsection{Extended Full Bayesian Model}
\label{sec:extensions}

The fully Bayesian model discussed in~\autoref{sec:full-model} requires additional information to be integrated as prior knowledge within the algorithm. In this section we present two extensions which leverage two different sources of expert knowledge: First we explain how to integrate feedback from human models, and afterwards we show how to use previous work on extracting sizing information from article images~\cite{sizenet} to incorporate prior information about articles even before any order is observed.
The combination of these extensions allows us to tackle the cold-start problem in a robust and efficient manner, constituting our final SizeFlags algorithm. Finally, we give an overview of the whole algorithm written as pseudo code.

\noindent\textbf{Human expert feedback:} To incorporate expert knowledge from human feedback we ask fashion models with relatively standard body sizes to try on articles before they are activated on the platform to give us early size and fit feedback. For the purpose of this paper, we constrain the feedback to be one of \{``certain size issue", ``potential size issue", ``good fit"\}. Multiple human models can fit the same article and based on the aggregated feedback we define article-specific prior parameters $(\alpha, \beta)$. If the aggregated feedback points strongly in the direction of a size issue, we chose values which strongly favor high return rates, if the aggregated feedback is more uncertain we favor more mildly high return rates, and if the aggregated feedback points towards a good fit we favor low return rates. 

\noindent\textbf{Size and fit visual cues:} Using the full Bayesian model presented in ~\autoref{sec:full-model} with human feedback does help alleviate the cold-start problem but it does not totally suppress the issue. In fact, a well trained group of human models can only fit a very limited part of the ever changing online assortment every week (less than $1\%$), which means only a few articles have prior information.
For our second extension, we thus propose to use size and fit visual cues to obtain prior information on large-scale and use it
in our full Bayesian model. Although the literature is very rich on computer vision techniques for various fashion problems, there are only few works considering fashion images for the size and fit problem (see \cite{hsiao20, sizenet} and references therein). We use our fully Bayesian model by incorporating sizing information extracted from article images using SizeNet~\cite{sizenet}. We use the student part in which a backbone convolutional neural network is used to extract visual features from article images and multi-layer perceptron learns article sizing behaviour. We convert the output of the network into prior values $(\alpha, \beta)$ of the Beta distribution. More specifically, the output is a size issue probability and, similarly to human expert feedback, we chose values which strongly favor high return rates for articles that have high size issue probability. 

To give the reader an overview of the SizeFlags algorithm, a summary in the form of a pseudo code is provided in Algorithm~\ref{algo-pseudo code}. 
\begin{algorithm} \caption{SizeFlags} \label{algo-pseudo code}
\begin{algorithmic}[1]
\Require Category $\mathcal{C}=\{a_1,\ldots,a_N\}$ contains $N$ articles $a_i$ with attributes \{$k_i$ size-related returns, $n_i$ number of orders\}
\State \textbf{Initialize} 
\begin{itemize}
    \item $srr(a_i) = k_i/n_i\, \forall a_i\in \mathcal{C}$
    \item $\pi= \text{mean}(\{srr(a_i)\}_{a_i\in \mathcal{C}})$
    \item $\sigma = \text{std}(\{srr(a_i)\}_{a_i\in \mathcal{C}})$
\end{itemize}
\ForEach {$a_i \in \mathcal{C} $}
  \If{$srr(a_i) \geq \pi + \sigma$}
     \State use prior information to set $(\alpha_i,\beta_i)=(\alpha(a_i),\beta(a_i))$
     \If{$- \log p(\pi| k_i,n_i; \, \alpha_i, \beta_i) > \theta$}
        \State $a_i$ has size issue: raise sizing flag
   \Else
     \State $a_i$ has probably no size issue: don't raise sizing flag
    \EndIf
   \Else
     \State $a_i$ has no size issue: don't raise sizing flag
    \EndIf
\EndFor
\end{algorithmic} 
\end{algorithm} 
\vspace*{-0.8\baselineskip}

\section{Experimental Results and Discussion}
\label{sec:experiments} 
\subsection{Challenges in Evaluating Size and Fit Recommendations}
Evaluating size and fit recommendations remains highly challenging due to multiple underlying factors specific to this problem space, in particular: 
\begin{enumerate}
\item Starting with the customer, the ``true'' size of a customer is often unknown, remains subjective for each customer, may depend on the context and the occasion for which a customer is shopping, and can vary greatly by external factors such as fashion trends, seasonality, and life changing events impacting a customer's physical body, and/or mindset around what fits best.

\item  Looking at the problem from the article side, the right size for a customer is not a unique quantity and varies greatly both within and across brands, in different sizing systems, in different countries, and for different fashion categories.

\item  In online fashion, there is a significant delay between the time that a size and fit recommendation is provided to a customer, and the feedback signal coming from the customer once they have actually tried the recommended size on. Through the lens of size and fit related returns, in order to evaluate the quality and effect of a recommendation, one needs to wait for several days if not weeks for the customer's return to reach the platform.

\item  From the customer experience point of view, customer satisfaction from a size and fit recommendation is not solely based on the recommendation's quality itself but is often intertwined with the physical characteristics and the experienced ``feel'' of the shoe or the garment once it is worn. This satisfaction is time-dependent and may change radically after wearing an article for a few weeks, or after a washing experience.
\end{enumerate}

Considering the above challenges, previous work~\cite{Sembium2018, guigoures18,julia20} has been mainly focused on evaluating the quality of the recommendations \textit{not} on their impact on reducing size-related returns, but rather on a combination of customer based metrics through A/B tests~\cite{Young2014} or continuous evaluation frameworks. Such customer based metrics include monitoring a change in customers' conversion rate, in the number of products added to the cart, in the revenues per visit, in selection orders (where a customer orders the same article in multiple sizes), and in reorders (where a customer returns an article and reorders it in a different size). In the same spirit, the literature also considers the customer's acceptance of the recommendations (i.e. how often they order in a recommended size), and accuracy of such recommendations (i.e. how often they keep or return a recommended size when they have accepted or not that recommendation). For more details on this direction, readers are invited to see~\cite{Sembium2018, guigoures18,julia20} and references therein. Within this work, we propose to go one critical step ahead and rigorously assess, for the first time to the best of our knowledge, the quality of the recommendations on their impact in reducing the size and fit returns.

In what follows, we first establish the state-of-the-art baseline and then evaluate our models, summarized in~\autoref{tab:versions}, from the binomial model launched for textile and shoe categories all the way to the full Bayesian SizeFlags launched in 2020 over 14 European countries with various local and mix size systems.
\begin{table}[t!]
    \centering
    \caption{Name abbreviations of the proposed models}
    \vspace*{-0.7\baselineskip}
    \begin{tabular}{ll}
         \hline
        $V_0$ & Binomial model on order and
         return data \\
                 $V_{HF}$ &  Only human feedback \\
        $V_{Base}$ & Baseline \\ & (order and return data + human feedback prior) \\
        $V_{SN}$ & Baseline + size and fit visual cues prior \\
        $V_{TH}$ & Baseline + optimized thresholds  \\
        SizeFlags & Baseline + size and fit visual cues prior \\ & \hspace{3.4em} + optimized thresholds \\ \hline\vspace*{-2.5em}
    \end{tabular}
    \label{tab:versions}
\end{table}
\subsection{Establishing state-of-the-art benchmark}

In online fashion reducing size-related returns is achieved through an algorithmically driven user experience, where different machine learning and recommendation approaches deliver a size advice to the customers and aim to drive their behaviour towards selecting the size that fits them the first time ~\cite{seewald2019towards}. Within this domain, controlled live A/B testings~\cite{Young2014} are widely used as a proven mechanism to benchmark the effect of such algorithmically driven user experiences on a set of metrics of interest. Therefore, to assess our approach we first establish our baseline to be the personalized size-recommendation in online fashion introduced in~\cite{guigoures18}, a solid state-of-the-art within this domain. Following~\cite{guigoures18} a customer receives a recommendation on which size to order for a given article based on their order and return history. Two successive live A/B tests were performed respectively on (1) women and men shoes and (2) women and men textile with over 300k customers per group. Using controlled and live A/B test settings, we randomly divided the customers into two groups; a control group that received no size advice and a test group with personalized advice delivered by the approach from~\cite{guigoures18}. The results showed positive financial impact with a significant increase in conversion rate (+2.1\%), more products added to the cart (+1.8\%), and increased revenues per visit (+2.1\%), however, no statistically significant impact on reducing size-related returns were observed (<0.5\% relative reduction).

\subsection{Evaluating our SizeFlags Models}

\textbf{A/B test on shoes ($V_0$):} 
To evaluated our binomial model (denoted $V_0$ in~\autoref{tab:versions}), we performed a first A/B test on women and men shoes in 2017. We provided the size advice derived from the size issue predictions of $V_0$ 
to the test group and no advice to the control group, both groups with 720k customers. The A/B test results demonstrated that $srr$ was significantly reduced by the provided size advice (3.8\% relative reduction). Interestingly, the A/B test also demonstrated that $V_0$ has a clear effect on size-related selection orders (customer orders two or more different sizes of the same article at the same time): the size advice `too small' led to a size-related selection order increase of 11.1\%, whereas `too big' led to an increase of 19.0\%, hinting at potentially different customer perception on whether a smaller or bigger size is a low risk order or flattering. 

\noindent\textbf{A/B test on textile  ($V_0$):} A second A/B test was performed for $V_0$ on textile in 2017. Each group with over 180k customers. In addition to all the usual textile categories such as dresses, trousers, etc. the sub-categories `beach \& lingerie' and `sportswear' were also included in this group. The A/B test results demonstrated that the size-related return rate was significantly reduced thanks to the provided size advice by $4.3\%$ for `too small' and $6.6\%$  for `too big' size flags. Interestingly, we observe that customers react stronger to the size advice for articles that are flagged as too big.

\subsection{Continuous evaluation}\label{subsec:comparable_groups_subsection} Outside of A/B testing, estimating the causal effects of size advice is a non-trivial exercise as very little is known about the individual customer behaviour before ordering an article with size advice. A/B tests provide an accurate snapshot of the intervention effect, however they are detrimental to customers' experience when continually performed since, by design, a group of customers end up with the relatively less attractive experience for the duration of the test. To tackle this challenge, here we first introduce an efficient approach for continuously monitoring the impact of our models (outside of A/B tests) and next present the experimental results of our models under the continuous evaluation regime. 

As discussed in \cite{heckman1997matching}, the nearest neighbor Difference-in-Diffe\-rences (DiD) matching estimator method is widely used in impact evaluation studies. We adopt DiD since it is able to compare the evolution over time $t$ of the size-related return rate $srr$ in two groups of treatment $\Omega_T$ and control $\Omega_C$, with parallel trends: $\Omega_T\subset \mathcal{C}$ that covers all articles with size advice in a given category $\mathcal{C}$ and 
$\Omega_C\subset \mathcal{C}$ that counterparts articles 
which have never had a size issue flag. From $\Omega_C$ we can find $\Omega_C^{\star}$,  with say the ten ($k=10$) nearest neighbors within the same category $\mathcal{C}$ of article $a$ where $a\in\Omega_T$ sold during the same time period. We use Euclidean distance over four continuous covariates to determine the nearest neighbours; (1) general return rate (including non size-related reasons), (2) price, (3) discount rate, and (4) return rate for the return type `unknown', as they have proven to be the four most common confounders in this context. Let's denote $srr(\Omega_{C}^{\star})$ as the average $srr$ of its elements and $t_{flag}(a)$ as the point of time when an article $a$ received a sizing flag. Using $\Omega_{C}^{\star}$ we are now able to estimate the average effect 
\vspace*{-0.7\baselineskip}

$$srr_{effect} = |\Omega_T|^{-1}\sum_{a \in \Omega_T}{\Big(\frac{\gamma_{a,{t_0}} - \gamma_{a,{t_1}}}{srr(a|t<t_{flag}(a))}}\Big)$$
\vspace*{-0.7\baselineskip}

of the size issue flags where
\begin{align*}
      \gamma_{a,{t_0}} = srr(a|t<t_{flag}(a)) - srr(\Omega_{C}^{\star}|t<t_{flag}(a)),\\
      \gamma_{a,{t_1}} = srr(a|t>t_{flag}(a)) - srr(\Omega_{C}^{\star}|t>t_{flag}(a)).
\end{align*}
In our estimations $t<t_{flag}(a)$ denotes a period of six weeks before the size issue flag was raised for an article $a$ and $t>t_{flag}(a)$  covers a period of six weeks after the size issue flag was raised.

\subsubsection{Impact of human expert feedback:}\label{sec:impact-hf} We focus here on the version $V_{Base}$ described in~\autoref{sec:extensions} which was launched in 2017 in order to tackle the cold-start problem by incorporating human expert size and fit feedback as a prior to our model. We utilize the continuous evaluation method DiD presented in~\autoref{subsec:comparable_groups_subsection} to estimate the impact of this prior on the size-related returns. To isolate and demonstrate the impact of this prior, we first focus on $V_{HF}$ flags which represent the subset of flags raised thanks to the human expert feedback ($V_{HF}\subset V_{Base}$). Afterwards, we will discuss the reached impact from the whole $V_{Base}$.

\noindent\textbf{Impact of $V_{HF}$:} 
\autoref{tab:human_expert} summarizes the strong positive impact of $V_{HF}$
in terms of srr reduction observed between September and December 2019. 
For the evaluation, 2678 articles were used that received a size advice too big or too small thanks to the human feedback prior. 
Our samples consist of 469 textile articles with 27773 orders and 4411 size-related returns and 2219 shoe articles with 53951 orders and 7846 size-related returns. As for the impact on the cold-start problem, it was observed that incorporating this prior leads to $33\%$ faster flagging, which entails that customers are informed about size issues much earlier, resulting in fewer size-related returns as shown in~\autoref{tab:human_expert}.

\vspace*{-0.7\baselineskip}
\begin{table}[!htbp]
\centering
\caption{Estimated impact of $V_{HF}$}
\vspace*{-0.8\baselineskip}

  \begin{tabulary}{\linewidth}{C|C|C}
    \mbox{category} & \mbox{srr reduction} & \mbox{avg. returns saved/article}  \\\hline
     Shoes & 4.40 \% & 5.68 \\
     Textile & 8.15 \% & 5.87 \\\hline
  \end{tabulary}  
  \label{tab:human_expert}
\end{table}
\vspace*{-0.5\baselineskip}

\noindent\textbf{Impact of $V_{Base}$:}
~\autoref{tab:DiD} summarizes the strong positive impact of $V_{Base}$
in terms of srr reduction observed between March 2019 and February 2020. The results are the average calculated effect for all articles $a\in\Omega_T$. 
The resulting group is much larger and consists of 10704 textile articles with 645667 orders and 120057 returns and 3625 shoes with 503849 orders and 80339 size-related-returns. The estimated impact also remains in line with the results obtained from our A/B tests.

\vspace*{-0.7\baselineskip}%
\begin{table}[!htbp]
\centering
\caption{Estimated impact of $V_{Base}$.} 
\vspace*{-0.8\baselineskip}

      \begin{tabulary}{\linewidth}{C|C|C}
    \mbox{category} & \mbox{srr reduction} & \mbox{avg. returns saved/article}  \\\hline
     Shoes & $5.65 \%$ & 2.29 \\
     Textile & $5.01 \%$ & 1.50 \\\hline
    \end{tabulary}
    \label{tab:DiD}
\end{table}%
\vspace*{-\baselineskip}

\subsubsection{Impact of SizeNet and threshold optimization:}\label{sec:impact-sn-th} 
\begin{figure}[ht!]
    \centering
    \includegraphics[width=\linewidth]{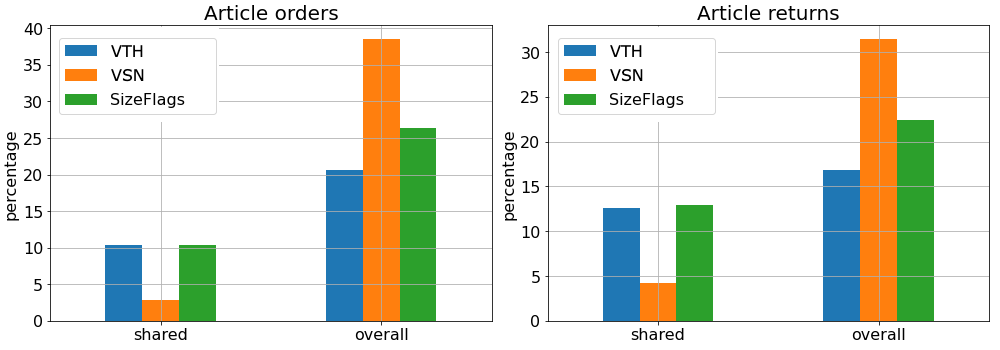}
    \vspace{-0.8cm}
    \caption{Impact of $V_{SN}$, $V_{TH}$, and SizeFlags compared to $V_{Base}$. Left: average relative reduction 
    of $n(a)$: the number of necessary article orders before a flag is raised. Right: average relative reduction of $ret(a)$: the number of necessary article returns before a flag is raised.}
    \Description{Impact of $V_{SN}$ and $V_{TH}$ compared to $V_{Base}$ reducing the average number of necessary article orders before a flag is raised. and the number of necessary article returns before a flag is raised.}
    \vspace*{-\baselineskip}
    \label{fig:time_to_raise}
\end{figure}

We evaluated the impact of  $V_{SN}$ and $V_{TH}$ along with the full SizeFlags model in comparison to the reference model $V_{Base}$. We ran these versions simultaneously over a 1 month period following the launch of the SizeFlags in 2020. The data for the simultaneous evaluation was limited to textile articles, the size of the dataset was 311851 articles and 25\% of them are flagged in the shared group. 
To assess the impact, we focus on the two key metrics illustrated in~\autoref{fig:time_to_raise}. (1) Article orders $n(a)$: the number of necessary article orders before a flag is raised, (2) Article returns $ret(a)$: the number of necessary article returns before a flag is raised. Since the evaluated versions $V_i$ may raise flags for articles which are not flagged by $V_{Base}$, 
we split our analysis between \textit{overall} articles $\Omega_O(V_i)$ and \textit{shared} articles $\Omega_S(V_i)$. $\Omega_O(V_i)$ refers to all the flags raised by $V_i$ and $\Omega_S(V_i)$ refers to those articles which are flagged by both $V_i$ and $V_{Base}$. 
Indeed, the articles in $\Omega_S(V_i)$ are particularly important as they have already proven to have good quality and impact. This leads us to set $\Omega_S(V_i)$ for all $V_i$ to cover at least $95\%$ of those raised by $V_{Base}$. To ensure quality, the articles in $\Omega_O(V_i)$ are continuously assessed following the approach DiD detailed in~\autoref{subsec:comparable_groups_subsection}. 
Based on~\autoref{fig:time_to_raise}, we observe that (1) both $V_{SN}$ and $V_{TH}$ positively reduce $n(a)$ and $ret(a)$ and (2) that $\Omega_S(V_{TH})$ outperforms $\Omega_S(V_{SN})$ whereas $\Omega_O(V_{SN})$  surpasses $\Omega_O(V_{TH})$. 
Since the flags raised by $V_{Base}$ 
have shown to bring strong positive impact on the $srr$ reduction, it is critical to guarantee that $\Omega_S(V_i)$ are raised even faster, and thus we opted for the balanced version SizeFlags in production which combines the two extensions $V_{SN}$ and $V_{TH}$, thereby bringing the best out of both of them to millions of customers. SizeFlags were rolled out to 14 countries in February 2020; since going live with this approach, we have seen the average of $n(a)$ and $ret(a)$ to drop significantly, by $17\%$ and $40\%$ respectfully, in the real world production environment.
\vspace*{-1\baselineskip}

\vspace*{-0.4\baselineskip}

\section{Conclusion}
\label{sec:conclusion}
\noindent The challenging task of reducing size and fit related returns in the context of online fashion was~studied.~A~probabilistic Bayesian approach (SizeFlags) was introduced leveraging large-scale return data and alleviating the cold-start problem thanks to rich priors from human experts feedback and computer vision techniques. The production results demonstrated that SizeFlags effectively reduces size-related returns in online fashion. In addition, using optimized thresholds and rich priors greatly reduced the number of ordered and returned articles necessary for reducing size-related returns. Limitations of our work include the coarse definition of size and fit issues (too small and too big), and future work will explore extending the Bayesian model to a hierarchical model with a multinomial likelihood and Dirichlet prior in order to include finer-grained size and fit problems (e.g. tight on the hips, too long sleeves) and more high fidelity data such as article measurements.
\section{Acknowledgements}
\label{sec:acknowledgements}

The authors would like to thank Romain~Guigourès and Yuen~King~Ho for the fun, positive energy, and  
fruitful size and fit discussions contributing to the success of this work.

\bibliographystyle{ACM-Reference-Format}
\bibliography{sample-base}

\end{document}